\documentclass[runningheads]{llncs}
    \usepackage{graphicx}
    \usepackage{bookmark}
    \usepackage{xcolor}
    \usepackage{verbatim}
    \usepackage[utf8]{inputenc}
    \usepackage{hyperref}
    \usepackage{amssymb}
    \usepackage{amsmath}
    \usepackage{stmaryrd}
    
    \usepackage{amsthm}
    \usepackage{csquotes}
    \usepackage{listings}
    \usepackage{nameref}
    \usepackage{svg}
    \usepackage{calc}
    \usepackage{tikz}
    \usepackage{newfloat}
    \usepackage{makecell}
    \usepackage{enumitem}
    \usepackage{subfig}
    \usepackage{multirow}
    \usepackage{float}
    \usepackage{booktabs}
    \usepackage{arydshln}
    \usepackage{todonotes}
    \usepackage{xcolor}
    \usepackage{paralist}
    
\restylefloat{table}
\usepackage{wasysym}

\newcommand{\mypar}[1]{
\smallskip\noindent\textbf{#1}
}

\begin{document}
\title{Formal Foundations of\\Agentic Business Process Management} 
% 
% \author{Anonymous submission}
% \institute{}
\author{Giuseppe De Giacomo\inst{1} \and Timotheus Kampik\inst{2,3} \and Lukas Kirchdorfer\inst{2,4} \and \\   Marco Montali\inst{5} \and Christoph Weinhuber\inst{1}}

\institute{
	University of Oxford, UK \\
    \email{firstname.lastname@cs.ox.ac.uk} 
    \and
    SAP, Germany \\
    \email{firstname.lastname@sap.com}
    \and
	Umeå University, Sweden
    \and
	University of Mannheim, Germany
    \and
    Free University of Bozen-Bolzano, Italy \\
    \email{marco.montali@unibz.it}
}
\authorrunning{De Giacomo et al.}

\titlerunning{Formal Foundations of Agentic BPM}
\maketitle
\begin{abstract}
Just like traditional BPM systems, agentic BPM systems are built around a specification of the process under consideration. Their distinguishing feature, however, is that the execution of the process is driven by multiple autonomous decision-makers, referred to as \emph{agents}. Since such agents cannot be fully controlled, the process specification is augmented with explicit objectives, or \emph{goals}, assigned to the participating agents. Agents then pursue these goals, at least to the best of their efforts, under suitable assumptions on the behavior of others, by adopting appropriate \emph{strategies}.
Centrally, the organization enacting the process can use these specifications to provide guardrails on the decision-making capabilities of agents at the strategy level.
This paper sets up the mathematical foundations of such systems in three key settings and analyzes four foundational problems of agentic BPM.
\keywords{Agentic BPM \and Goals \and Autonomous Agents.}
\end{abstract}

%
%
%%%%%%%%%%%%%%%%%%%%%%
\section{Introduction}
\label{sec:intro}
%%%%%%%%%%%%%%%%%%%%%%
%\todo[inline]{
%Some important remarks. Distribution for us is distribution of decision making. We do not care about communication protocols and distribution of message exchange. 
%This is also why distribution boils down to centralized if all agents are cooperative and their goals are consistent.
%}
A business
process is commonly defined as a ``set of activities that are performed in coordination'' and that ``jointly realize a business goal''~\cite[p.15]{weske2024business}.
Formal foundations for process science have long supported reasoning about processes and their executions using well-established frameworks such as Petri nets~\cite{10.1007/978-3-642-38143-0_4} and linear temporal logic (LTL) on finite traces~\cite{DegVa13,DBLP:books/sp/22/CiccioM22}.
%
%These foundations model processes as logically connected \emph{sets of activities}, while giving much less attention to \emph{coordination}, \emph{goals}, and their mutual interaction.
These foundations model processes as logically connected \emph{sets of activities}, while giving much less attention to \emph{goals} that are pursued by agents and their mutual interaction.

Process models and corresponding analysis and mining techniques typically target control flow, possibly enriched with additional perspectives, but do not explicitly account for goals. Goals are compiled away into designated final states in the process model, such as multiple end events in a BPMN diagram or final markings in an accepting Petri net. 
In related fields such as requirements engineering, goals have been studied in depth and are represented through dedicated models, such as $i^*$ diagrams. 
However, the integration between goals and processes is mostly confined to modeling and analysis~\cite{DBLP:journals/is/RuizCEFP15,MTZM11}, with little consideration on how goals influence decision-making at runtime. 
Approaches that make goals explicit, such as fragment-based case management~\cite{DBLP:conf/bis/HaarmannW20}, use them primarily as acceptance conditions on executions rather than as drivers of strategic decision-making during enactment.
%
%When one is primarily interested in the organization-level view of a process, the simplified representation with implicit goals (which may, for example, be assumed to be reached when a token is in the final place of a sound workflow net) is intuitive. %, and can be augmented by measuring process performance indicators such as cycle time and costs that aggregate attributes on process instance and activity level.
%

Linking goals to runtime decisions becomes even more important when tackling the \emph{coordinated execution} of activities, which hints at the presence of multiple actors, resources, or \emph{agents} within the process. Multi-party processes have been extensively studied, %also from the foundational point of view:
from process choreographies~\cite{DBLP:conf/bpm/BarrosDO05} and techniques for deriving (stubs of) specifications for single participants \cite{DBLP:journals/ipl/MassutheSSW08,DBLP:journals/is/AalstLR12},  to declarative process models lifting well-known approaches such as DCR Graphs and Declare to multi-party, collaborative, and inter-organizational processes \cite{Hildebrandt2019,GeattiMR23}.
%augments Declare with controllable (internal) actors and uncontrollable (external) actors.
%In this vein, interaction protocols provide a declarative counterpart to (usually procedural) choreography specifications~\cite{DBLP:journals/tse/DesaiMCS05}.

%What is not explored in these works is the intimate connection between coordination and goals, a cornerstones of multi-agent systems in AI \cite{wooldridge2009mas}. It is in fact natural to equip each agent with dedicated goals, so as to specify what the agent should achieve, leaving the agent free to autonomously decide how to. The issue is that since an agent controls only a portion of the overall behavior, coordination with the other agents is essential to reach the goal, while ensuring that the overall process is executed properly. This calls for developing \emph{formal foundations of agentic BPM}: process-aware, formal representations and corresponding reasoning techniques for agents that (inter)act \emph{autonomously} and \emph{strategically} when jointly executing a process. 

Agents are naturally modeled as autonomous entities with individual goals, which determine what should be achieved while leaving open how to act. Since an agent controls only a portion of the overall system, coordination is an essential tool to enable achieving the agent's goals. In a BPM context, the additional complication is that agents should also guarantee to stay within the boundaries of the process they are co-enacting.
This calls for formal foundations for \emph{agentic BPM}: process-aware models and reasoning techniques for autonomous, strategic agents that jointly enact a business process.

%In existing approaches to process specifications, agent-like notions occur in several forms.
%In BPM, work is typically assumed to be executed by \emph{resources} (e.g., \emph{swimlanes} in BPMN).
%More fine-grained methods for differentiating between the type of involvement of resources in the execution of activities exists in the form of so-called \emph{responsibility assignment matrices}~\cite{DBLP:conf/otm/CabanillasRC12}.
%\lnote{Can we shorten the whole part on mining/discovery-based approaches?}

% Related ideas also appear in artifact-centric collaborative processes~\cite{EshuisHSV13}. 
% \lnote{Maybe drop next paragraph}
% While these approaches move beyond monolithic execution and account for different sources of control, their main focus is on interaction structure, realizability, and orchestration under assumptions about external behavior. In contrast, our work explicitly models autonomous agents through individual goals and reasons directly over the resulting strategy spaces.

%In other contexts, however, goals may play a crucial role: organizations are often not intrinsically process-aware, but rather rely on document-oriented IT systems, as well as on somewhat autonomous \emph{agents} to achieve business goals.

\begin{figure}[t]
    \centering
    \includegraphics[width=.8\linewidth]{figures/tree.pdf}
    \caption{Agentic BPM is structured along three dimensions: (i) decision-making (centralized vs. decentralized), (ii) agents’ assumptions about peers (none/explicit/implicit), and (iii) goal attainment (winning vs. best effort).}
    \label{fig:tree}
    \vspace{-1.5em}
\end{figure}

In this paper, we develop these foundations. 
We study goal-oriented agents jointly executing a process along the three dimensions of Figure~\ref{fig:tree}: centralized or decentralized decision-making, different assumptions about the behavior of other agents, and different notions of goal attainment. In this setup, an agent must choose which action to execute next based on the history so far, its own goals, the process boundaries, and the possible behavior of others.
%
%We consider goal-oriented agents jointly executing a process, studying different settings depending on the three main dimensions shown in Figure~\ref{fig:tree}. 
%Agents may cooperate and compete, with or without assumptions on what the other agents do, and with the intention to strongly achieve their goals, or trying to find the best compromise. They thus need to devise execution strategies to achieve their goals depending on their own previous choices, and on what the other agents do. A \emph{strategy} is in fact a function that describes which action the agent takes given the history of observations so far. A strategy is not only conditioned by the agent's goal and by the behavior of other agents, but also by the boundaries dictated by the process. This results in the formation of \emph{guardrails}, which capture the (infinite) set of strategies that are allowed for each agent guaranteeing that if that agent operates within that set, the process is enacted properly and the goal is achieved, at least to the best of the agent's effort. We formalize the dimensions from Figure~\ref{fig:tree}, resulting in three main configurations of an agentic BPM system. For each such configuration, we characterize complexity and algorithmic solutions for key problems related to the existence and synthesis of strategies.
%
We formalize such choices as \emph{strategies}, and characterize for each agent a \emph{guardrail}: the (typically infinite) sets of strategies that are admissible with respect to process constraints and goal satisfaction.
%We refer to these sets as \emph{guardrails}.
Building on this formalization, we identify three main settings of agentic BPM systems, and for each of them study key problems related to the existence and synthesis of strategies and guardrails.
Our results establish complexity bounds and algorithmic techniques to tackle these problems.

More broadly, our framework makes goals and strategies first-class abstractions in BPM. 
In this respect, it provides a formal basis for the emerging discipline of \emph{agentic business process management}, in which software agents---typically based on generative AI and large language models---are granted controlled autonomy in the execution of organizational work~\cite{10.1007/978-3-032-02936-2_3,calvanese2025autonomy}.

%Our foundations bring agents' goals and strategies to first-class abstractions for representing and reasoning about process specifications. In this respect, they provide a formal point of departure for the emerging discipline of \emph{agentic (business) process management}, in which software agents---often based on large language models---are granted some degree of autonomy when conducting work in organizations~\cite{10.1007/978-3-032-02936-2_3,calvanese2025autonomy}.

%In this write-up, we sketch how to represent goals of agents in business processes such that set of strategies can be inferred from these goals and (in our case \emph{declarative}) operational process specifications.
The paper is structured as follows. We first provide a motivating example illustrating why representing the goals of agents and reasoning about agents' strategies is important in BPM (Section~\ref{sec:motivation}).
%Based on preliminaries from presented in Section~\ref{sec:prelims},
We then introduce the mathematical foundations of agents in business processes (Section~\ref{sec:formal}), based on which we define three key \emph{settings} for agents in BPM (Section~\ref{sec:settings}).
We introduce four computational problems that are important for agentic BPM (Section~\ref{sec:services}) and analyze their computational properties, given the three settings and assuming process specifications in Linear Temporal Logic (LTL) (Section~\ref{sec:ltl_instantiation}).
Finally, we conclude with a future outlook (Section~\ref{sec:discussion}). %before we conclude with a brief future outlook (Section~\ref{sec:concl}).

\section{Motivating Example}
\label{sec:motivation}
\begin{figure}[t]
    \centering
        \includegraphics[width=1\textwidth,clip,trim=185 345 185 100]{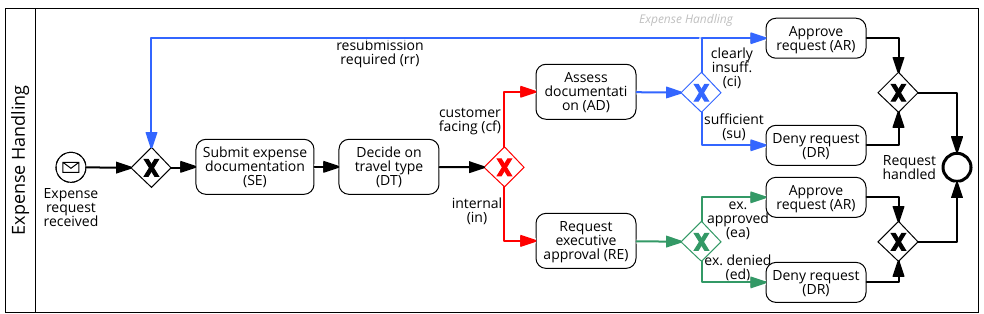}
\caption{The travel expense process, executed by three agents (indicated by the colored splitting XOR gateways) in a potentially adversarial environment.}
\label{fig:intro-agentic}
\vspace{-1.5em}
\end{figure}

% \lnote{Add note on imperative vs. declarative (see comment in discussion) -> BPMN is just an intuitive way to describe}
Consider a travel expense process. A traveler submits an expense claim and marks it as either \textit{customer-facing} or \textit{internal}. Customer-facing expenses are assessed for documentation completeness by an expense handler, whereas internal expenses require approval by the manager. 
For an intuitive understanding, we model the process as a \emph{Business Process Model and Notation} (BPMN) diagram (cf. Figure~\ref{fig:intro-agentic}).
From a traditional BPM perspective, this process can be executed automatically: the traveler provides attributes (travel type and documentation), and rule-based decisions route the case through the appropriate path.

% \begin{figure}[t]
%     \centering
%         \includegraphics[width=1\textwidth,clip]{figures/process_agentic.png}
% \caption{The travel expense process, executed by three agents (indicated by the colored splitting XOR gateways) in a potentially adversarial environment.}
% \label{fig:intro-agentic}
% \vspace{-1.5em}
% \end{figure}

However, even simple operational processes such as travel expense reimbursement involve \emph{agents} that act and take decisions driven by goals. Making these goals explicit reveals important nuances that are overlooked in traditional BPM. 

\mypar{A brief note on LTL.}
In the remainder of this section, we present  an intuitive description and a formal representation, using LTL\footnote{We use LTL rather than its finite trace variant LTL$_f$ more commonly used in BPM, because we rely on existing results on multi-agent strategic reasoning \cite{KR2025-69,DBLP:conf/stacs/Berwanger07}, which have been developed for LTL but not yet for LTL$_f$. Note that every LTL$_f$ formula can be translated into an equivalent LTL formula in polynomial time~\cite{DBLP:conf/aaai/GiacomoMM14}, so our results apply to LTL$_f$ as well. However, the constructions could be considerably simplified for LTL$_f$. We leave this for future work.}, 
of the example process from an agent-oriented perspective. 
Note that a detailed understanding of LTL is not required before Section~\ref{sec:ltl_instantiation}; this section can therefore also be read on the intuitive level. 
Further, our modeling of agentic processes is largely agnostic to the underlying paradigm, supporting both imperative (e.g., Petri net) and declarative (e.g., LTL) approaches. In this paper, however, we use LTL, also because it supports our later complexity analysis.
We refer to~\cite{BaierK08} for the formal definition of LTL. For now, it is sufficient to read the temporal operators informally: $\Diamond \varphi$ means that $\varphi$ holds eventually, $\Box \varphi$ means that $\varphi$ holds always, and $\Circle \varphi$ means that $\varphi$ holds at the next step. 
Likewise, $\varphi \,\mathcal{U}\, \psi$ means that $\varphi$ holds until $\psi$ holds,
and $\varphi \,\mathcal{W}\, \psi$ is the weak-until variant, where we do not require $\psi$ to ever hold.

%\begin{description}[style=unboxed,leftmargin=0cm]

\mypar{Process specification, agents, and goals.}
    As shown in Figure~\ref{fig:intro-agentic}, the process consists of seven unique events\footnote{In case of activities, we consider for simplicity events witnessing their completion.}: $\{SE,DT,AD,RE,DR,AR,End \}$ (considering an explicit end event). We model these events and the associated data attributes, which in this example are limited to the documentation quality ($DG$), as fluents. The process involves three actors (i.e., agents) with different goals: \textbf{traveler}, \textbf{expense handler}, and \textbf{executive manager}. The agents' actions correspond to the outgoing branches of the XOR gateways, i.e., $\{cf,in,ci,rr,su,ea,ed \}$.

    After the start event, the process always begins with submitting the expense documentation (yielding the following LTL formula: $\varphi_1 = SE$). This event is always followed by the decision on the travel type ($\varphi_2 = \Box(SE \Rightarrow \Circle DT)$). 
    
    Now, \textbf{traveler} selects either customer facing ($cf$) or internal ($in$) as travel type. If internal, the next event is requesting executive approval ($\varphi_3 = \Box(DT \land in \Rightarrow \Circle RE)$), otherwise a documentation assessment follows ($\varphi_4 = \Box(DT \land cf \Rightarrow \Circle AD)$). \textit{Traveler's goal is to obtain reimbursement's approval} ($G_1 := \Diamond AR$).
    
    When the process is in the internal branch, the \textbf{manager} can either approve or deny the request ($\varphi_5 = \Box(RE \land ea \Rightarrow \Circle AR)$ and $\varphi_6 = \Box(RE \land ed \Rightarrow \Circle DR)$). \textit{The manager's goal is left unconstrained to eventually reach either decision outcome} ($G_3 := \Diamond(AR \lor DR)$). The idea is that its decisions should not be governed, as it is trusted to make better decisions based on informal knowledge.

    If the process is in the customer facing branch, the \textbf{expense handler} assesses documentation quality, deciding whether a resubmission is required ($rr$) by the traveler, which brings the process back to $SE$ (yielding $\varphi_7 = \Box(AD \land rr \Rightarrow \Circle SE)$), or whether the documentation is clearly insufficient ($ci$) (yielding $\varphi_8 = \Box(AD \land ci \Rightarrow \Circle DR)$) or sufficient ($su$) (yielding $\varphi_9 =  \Box(AD \land su \Rightarrow \Circle AR)$).
    \textit{The expense handler's goal is to get a request with good documentation accepted and to either get the request denied or resubmitted, otherwise} ($G_2 := \Box(DG \land AD \Rightarrow \Diamond AR)\ \land\ \Box(\neg DG \land AD \Rightarrow \Diamond(DR \lor SE))$).

    After reaching either approval or denial, the process reaches its end ($\varphi_{10} = \Box((AR \lor DR) \Rightarrow \Circle End)$), and once the end is reached, the process remains there ($\varphi_{11} = \Box(End \Rightarrow \Circle End)$).
    Furthermore, we say that if no valid action is taken at a decision point, the process remains in its current state (yielding $\varphi_{12} = \Box(DT \land \lnot(in \lor cf) \Rightarrow \Circle DT)$, $\varphi_{13} = \Box(AD \land \lnot(rr \lor su \lor ci) \Rightarrow \Circle AD)$, and $\varphi_{14} = \Box(RE \land \lnot(ea \lor ed) \Rightarrow \Circle RE)$).
    %Importantly, at every point in time, exactly one fluent holds ($\varphi_{15} = \Box(SE \oplus DT \oplus AD \oplus RE \oplus DR \oplus AR \oplus End)$).
    Importantly, at every point in time, exactly one event fluent holds ($\varphi_{15} = \Box((SE \lor DT \lor AD \lor RE \lor DR \lor AR \lor End) \land \bigwedge_{\{{\cal F}, {\cal F'} \in \{SE, DT, AD, RE, DR, AR, End\} | {\cal F} \neq {\cal F}'\}} \neg ({\cal F} \land {\cal F'}))$).
    %Also, if no action occurs, any fluent $\mathcal{F}$ persists (yielding $\varphi_{16} = \Box((\mathcal{F} \land \lnot(rr \oplus cf \oplus in \oplus ci \oplus su \oplus ea \oplus ed)) \Rightarrow \Circle \mathcal{F})$).
    Also, if no action occurs, any fluent $\mathcal{F}$ persists (yielding $\varphi_{16} = \Box((\mathcal{F} \land \lnot(rr \lor cf \lor in \lor ci \lor su \lor ea \lor ed)) \Rightarrow \Circle \mathcal{F})$).
    % \todo[inline]{$F \land (anything) \implies F$ is always true. I think we want to say: $F \land (anything) \implies \Circle F$ ?}
    Finally, the documentation quality remains as-is (weakly) until submission or resubmission ($SE$) occurs (yielding $\varphi_{17} = \Box(DG \Rightarrow \Circle(DG \,\mathcal{W}\, SE))$ and $\varphi_{18} = \Box(\neg DG \Rightarrow \Circle(\neg DG \,\mathcal{W}\, SE))$).

    Formally, the whole process specification is then described by $\Gamma = \bigwedge_{i=1}^{18} \varphi_i$.

    \mypar{Strategies.}
    Agents can thus be viewed as active decision-makers within a process, guided by their individual goals. Their behavior can be described in terms of \textit{strategies} that specify how an agent acts after any given trace of the process so far. For example, after submitting an expense request, the traveler can classify it as either customer-facing or internal. Since the process itself cannot verify this classification, the decision effectively rests with the traveler. From a self-interested perspective, the traveler may reason strategically about which option maximizes the likelihood of approval. In the case of good documentation, classifying the expense as $cf$ is the more favorable choice, given that the expense handler aims to approve well-documented requests. Viewing processes in this way shifts attention from predefined control-flow to the space of possible behaviors that autonomous agents may adopt within the allowed specification.

%\end{description}

\noindent 
Given explicit agent goals, different system designs become possible, which mirror classical distinctions in multi-agent system literature~\cite{wooldridge2009mas}.
A process designer may retain \emph{central control} (as in traditional non-agentic BPM), governing all decision points by specifying each agent's strategy s.t. all agents’ goals are satisfied whenever possible. Here, agent autonomy plays no role.

Alternatively, the designer may allow \emph{decentralized decision-making}, where each agent autonomously selects its strategy. In this case, the system must specify how agents reason about one another. One option is to provide \emph{explicit assumptions} about peers’ behavior---for instance, that the traveler assumes that the expense handler always requests resubmission for bad documentation. Another option is to leave such assumptions implicit and not prescribe how agents expect others to behave. 
Instead, it only requires that each agent believes that every other agent acts \emph{rationally}, in the sense of choosing strategies that aim to achieve their own goals within the process constraints.

On a high-level, this gives rise to three different settings of agentic BPM: (1) centralized, (2) decentralized with explicit assumptions, and (3) decentralized with implicit assumptions.
These settings correspond to fundamentally different implementations of agentic BPM, which we will formalize in Section~\ref{sec:settings} based on the foundations introduced next.

\section{Mathematical Foundations of Agentic Processes}
\label{sec:formal}

In this section, we introduce the foundations of agentic business processes in an entirely abstract, set-based manner.

\begin{definition}[Traces]
Let $\mathfrak{A}$ be a finite set of propositional variables, called \emph{local properties}.  
An \emph{instant} is a propositional interpretation of $\mathfrak{A}$, i.e., a subset of $\mathfrak{A}$.  
A \emph{trace} is an infinite sequence of instants, i.e., an element of $(2^{\mathfrak{A}})^\omega$, where $2^{\mathfrak{A}}$ denotes the set of all subsets of $\mathfrak{A}$ and $\omega$ indicates infinite repetition.  
A \emph{specification} identifies a subset of admissible traces from $(2^{\mathfrak{A}})^\omega$.
\end{definition}

\noindent Intuitively, a local property may represent, for example, that a certain activity has occurred or that a data attribute has a given value. A trace, therefore, captures the evolution of such properties over time, and a specification characterizes the executions permitted by a process.
For example, in the travel expense process, we partition the set of properties as
$\mathfrak{A} = Z \cup X_1 \cup \dots \cup X_n$,
where each $X_i$ represents the action space under the exclusive control of agent $\mathcal{A}_i$ where $i \in \{1,\dots,n\}$, and $Z$ contains the remaining fluents determined by the process specification.
% , which are $\{SE, DT, AD, RE, DR, AR, DG\}$.  
At each step of execution, agent $\mathcal{A}_i$ selects a valuation of the variables in $X_i$, thereby choosing one of its available actions.
For instance, in our running example, the agent-controlled variables are
$X_1 = \{cf, in\}, X_2 = \{ci, rr, su\}, X_3 = \{ea, ed\}$.
$Z$ instead is controlled by the environment, which determines the fluents when these are not fully determined by the agents' actions, e.g., the quality of the documentation of the submission ($DG$).

To capture decision-making by agents, we introduce the notion of a strategy.

\begin{definition}[Agent strategies]
Let $i \in \{1,\dots,n\}$.  
A strategy for agent $\mathcal{A}_i$ is a function
$\sigma_i : \mathfrak{A}^* \rightarrow 2^{X_i}$
that maps a finite execution history to the action chosen by agent $\mathcal{A}_i$ at the next step, i.e., the propositional evaluation of its variables $X_i$.  
Without loss of generality, we assume that at every moment an agent may do nothing, i.e., it performs a no-op action, which must be included in $2^{X_i}$. We use a special symbol $\bot$ to denote such action.
\end{definition}

Given a strategy $\sigma_i$, the agent selects at each step a valuation of its action variables $X_i$ based on the execution history so far. We assume perfect information, i.e., strategies observe the full trace.
Formally, the environment also adopts strategies, which are functions $\sigma_e : \mathfrak{A}^* \rightarrow 2^Z$. 
The joint execution of all agents’ strategies and the environment strategy induces a unique trace, denoted \emph{play}.

\begin{definition}[Play]
Let $\sigma_1,\dots,\sigma_n$ be strategies for agents $\mathcal{A}_1,\dots,\mathcal{A}_n$ and $\sigma_e$ be the strategy of the environment.  
The \emph{play} induced by $\sigma_1,\dots,\sigma_n, \sigma_e$, denoted
$\textit{play}(\sigma_1,\dots,\sigma_n, \sigma_e)$,
is the unique trace $t \in (2^{\mathfrak{A}})^\omega$ such that for all positions $k \ge 0$ and all agents $\mathcal{A}_i, i \in \{1,\dots,n\}$ and environment, it holds
$t[k]|{X_i} = \sigma_i\!\big(t[0,k-1]\big)$ and $t[k]|{Z} = \sigma_e\!\big(t[0,k-1]\big)$,
where $t[0,-1] = \epsilon$ (the empty trace).  
\end{definition}

That is, at each step agents choose their actions according to their strategies, and the fluent variables are uniquely determined by the environment strategy, given the chosen actions and the execution prefix.

\begin{definition}[Operational Process Specification]
\label{def:gamma}
Let $\Gamma \subseteq (2^{\mathfrak{A}})^\omega$ be the operational process specification. We denote by $\llbracket \Gamma \rrbracket$ the set of strategies that are compliant with $\Gamma$, that is 
$\llbracket \Gamma \rrbracket
\;=\;
\{\sigma_e \,|\,
\forall \sigma_1,\dots,\sigma_n.\ 
\textit{play}(\sigma_1,\dots,\sigma_n,\sigma_e)\in \Gamma
\}$.
We assume that $\Gamma$ is ``consistent'', that is $\llbracket \Gamma \rrbracket \neq \emptyset$.  
\end{definition}
% \lnote{Now we also have goals in the centralized setting?}
% In decentralized agentic BPM, the process does not prescribe which strategies agents must follow. Instead, it specifies for each agent which executions are desirable, i.e., its goal. These goals are expressed as subsets of $\mathfrak{A}^*$ characterizing the traces that are considered successful for the agent.

In agentic BPM, agents are guided by their individual goals. These goals are expressed as subsets of $(2^{\mathfrak{A}})^\omega$ characterizing the traces that are considered successful for the agent.

\begin{definition}[Agentic Process]\label{def:agentic-process}
An \emph{agentic process} is a tuple 
$(\Gamma,\mathcal{A},\mathcal{G})$,
where $\Gamma \subseteq (2^{\mathfrak{A}})^\omega$ is the operational process specification, 
$\mathcal{A} = \{\mathcal{A}_1, \dots, \mathcal{A}_n\}$ is a set of agents, each with decision-making capability, i.e., a set of strategies $\Sigma_i$ that it can choose from,
and $\mathcal{G}=\{G_1,\dots,G_n\}$ is a set of agent goals with $G_i\subseteq (2^{\mathfrak{A}})^\omega$ specifying the traces desirable for agent $\mathcal{A}_i$.
\end{definition}

When we do not specify the set of agent $\mathcal{A}_i$'s strategies $\Sigma_i$ explicitly, as in the next few paragraphs, we assume that they include all possible strategies. Given an agentic process $(\Gamma,\mathcal{A},\mathcal{G})$,  
a \emph{strategy profile} is a tuple of strategies 
$\upsilon = \sigma_1,\dots,\sigma_n$, %($\in \Sigma_1\times \cdots\times\Sigma_n$),
where $\sigma_i$ is a strategy of $\mathcal{A}_i$.  
For $i\in\{1,\dots,n\}$, we denote by 
$\upsilon_{-i} = \sigma_1,\dots,\sigma_{i-1},\sigma_{i+1},\dots,\sigma_n$
the profile of all strategies except that of $\mathcal{A}_i$.

\begin{definition}[Enforce for Strategy Profiles]
For a strategy profile $\upsilon = \sigma_1, \ldots, \sigma_n$ and a goal $G= G_1 \land \ldots \land G_n$ under the environment $\Gamma$, we define the set of strategy profiles that enforce $G$ under the environment $\Gamma$ as
\[
\text{ENF}(G)
\;=\;
\Big\{\upsilon  \,\Big|\,
\forall \sigma_e \in \llbracket \Gamma \rrbracket.\
\textit{play}(\upsilon,\sigma_e)\in G
\Big\}.
\]
\end{definition}

However, goals may not be fully enforceable, i.e., goal attainment may not be guaranteed.  
We therefore consider strategies that maximize goal achievement relative to others.
For this, we make use of the notion of best effort~\cite{DBLP:conf/ijcai/AminofGR21}.
\begin{definition}[Strategy Profile Dominance]
Let $G_i \subseteq (2^{\mathfrak{A}})^\omega$ be the goal for agent $\mathcal{A}_i$. A strategy profile $\upsilon$ \emph{dominates} another profile $\upsilon'$ for $G = G_1 \land \ldots \land G_n$ under the process specification $\Gamma$, denoted $\upsilon \ge_{G} \upsilon'$, iff for every environment strategy $\sigma_e \in \llbracket \Gamma \rrbracket$, the following holds:
$\textit{play}(\upsilon', \sigma_e) \in G \implies \textit{play}(\upsilon, \sigma_e) \in G$.
Furthermore, $\upsilon$ \emph{strictly dominates} $\upsilon'$, denoted $\upsilon >_{G} \upsilon'$, iff $\upsilon \ge_{G} \upsilon'$ and $\upsilon' \not\ge_{G} \upsilon$. 
\end{definition}

Based on the above definition of strategy profile dominance, we can now define the set of best-effort strategy profiles.
% \tnote{Someone add some ``glue'' between the definitions :)}

\begin{definition}[Best-Effort for Strategy Profiles]
A strategy profile $\upsilon$ is \emph{maximal}, or \emph{best-effort}, for the goal $G$ under the process specification $\Gamma$, iff there is no strategy profile $\upsilon'$ such that $\upsilon' >_{G} \upsilon$.
For a strategy profile $\upsilon = \sigma_1,\ldots,\sigma_n$ and a goal $G=G_1\land\ldots\land G_n$ under environment $\Gamma$, we denote the set of best-effort strategy profiles
$\text{BE}(G)
= \{
\upsilon
\mid 
% \forall \upsilon' 
% .\ \upsilon \geq_{G} \upsilon'
\neg \exists \upsilon' .\ \upsilon' >_{G} \upsilon
\}$.

% \[
% \text{BE}(G)
% \;=\;
% \Big\{\upsilon \,\Big | \, \forall \upsilon' 
% % \in \Sigma_1 \times \ldots \times \Sigma_n
% .\ \upsilon \geq_{G} \upsilon'
% \Big\}.
% \]

\end{definition}

In the above definitions, we are considering strategy profiles collectively without allowing that agents make decisions independently. To support this, we need to refine the above notions with individual agents in mind, thus focusing on individual strategy sets $\Sigma_i$ per agent $\mathcal{A}_i$.
A set of strategy profiles $\Sigma$ is \emph{rectangular} iff 
$\Sigma=\Sigma_1\times\cdots\times\Sigma_n$.
% where each $\Sigma_i$ is a set of strategies for agent $\mathcal{A}_i$.
We use $\Sigma_{-i} = \Sigma_1\times\cdots\times\Sigma_{i-1}\times\Sigma_{i+1}\times\dots\times\Sigma_n$ to denote the set of strategy profiles of all agents except $\mathcal{A}_i$.
If $\Sigma$ is rectangular, each agent can choose its own strategy $\sigma_i \in \Sigma_i$ independently of all other agents.  
\textit{In this work, we focus on rectangular strategy sets}.

\begin{definition}[Enforce Strategy]
For every $i\in\{1,\dots,n\}$, we define the set of strategies for agent $\mathcal{A}_i$ that \emph{enforce} $G_i$ under the environment $\Gamma$ and strategies $\Sigma_{-i}$ of the other agents as
\[
\text{ENF}_i(G_i,\Sigma_{-i})
\;=\;
\Big\{\sigma_i \,\Big|\,
\forall \sigma_e \in \llbracket \Gamma \rrbracket.\
\forall \sigma_{-i} \in \Sigma_{-i}
.\
\textit{play}(\sigma_i,\sigma_{-i},\sigma_e)\in G_i
\Big\}.
\]
\end{definition}

Intuitively, $\text{ENF}_i(G_i,\Sigma_{-i})$ is the set of strategies of agent $i$ that enforce its goal $G_i$ in $\Gamma$ \textit{provided} that every other agent $j \neq i$ uses a strategy from $\Sigma_{-i}$.

\begin{definition}[Strategy Dominance]
Let $\mathcal{A}_1, \dots, \mathcal{A}_n$ be agents with an individual goal $G_i \subseteq (2^{\mathfrak{A}})^\omega$, and $\Sigma_{-i}$ be a set of strategy profiles for the remaining agents. 
A strategy $\sigma_i$ \emph{dominates} another strategy $\sigma'_i$ for $G_i$, denoted $\sigma_i \ge_{G_i | \Sigma_{-i}} \sigma'_i$, iff for every profile $\sigma_{-i} \in \Sigma_{-i}$ and for every environment strategy $\sigma_e \in \llbracket \Gamma \rrbracket$, the following holds:
$\textit{play}(\sigma'_i, \sigma_{-i}, \sigma_e) \in G_i \implies \textit{play}(\sigma_i, \sigma_{-i}, \sigma_e) \in G_i$.
Furthermore, $\sigma_i$ \emph{strictly dominates} $\sigma'_i$, denoted $\sigma_i >_{G_i|\Sigma_{-i}} \sigma'_i$, iff $\sigma_i \ge_{G_i|\Sigma_{-i}} \sigma'_i$ and $\sigma'_i \not\ge_{G_i|\Sigma_{-i}} \sigma_i$.

\end{definition}

Given strategy dominance, we can define the set of best-effort strategies. % can be defined as follows.
% \tnote{As above, consider adding some ``glue''}

\begin{definition}[Best-Effort Strategy]
\label{def:best-effort}
An individual strategy $\sigma_i$ is \emph{maximal}, or \emph{best-effort}, for the goal $G_i$ under the process specification $\Gamma$ and peer strategies $\Sigma_{-i}$, iff there is no strategy $\sigma'_i$ such that $\sigma'_i >_{G_i|\Sigma_{-i}} \sigma_i$.

We denote the set of best-effort strategies for agent $\mathcal{A}_i$ and goal $G_i$ as
\[
\text{BE}_i(G_i, \Sigma_{-i})
\;=\;
\Big\{
\sigma_i \in \Sigma_i \,\Big| \,
% \forall \sigma'_i \in \Sigma_i.\ 
% \sigma_i \geq_{G_i|\Sigma_{-i}} \sigma'_i
\neg \exists \sigma_i' \in \Sigma_i .\ \sigma_i' >_{G_i|\Sigma_{-i}} \sigma_i 
\Big\}.
\]
\end{definition}

% \begin{definition}[Best-Effort Strategy]
% \label{def:best-effort}
% An individual strategy $\sigma_i$ is \emph{maximal}, or \emph{best-effort}, for the goal $G_i$ under the process specification $\Gamma$ and peer strategies $\Sigma_{-i}$, iff there is no strategy $\sigma'_i$ such that $\sigma'_i >_{G_i|\Sigma_{-i}} \sigma_i$.

% We denote the set of best-effort strategies for agent $\mathcal{A}_i$ and goal $G_i$ as
% \[
% \text{BE}_i(G_i)
% \;=\;
% \Big\{
% \sigma_i \,\Big| \,
% \forall \sigma'_i \in \Sigma_i.\ 
% % \forall \sigma_{-i} \in \Sigma_{-i}.\ 
% \sigma_i \geq_{G_i|\Sigma_{-i}} \sigma'_i
% \Big\}.
% \]

% Similarly, we can introduce the set of best-effort strategies for agent $\mathcal{A}_i$, goal $G_i$ and assumptions $\Theta_i$ as $\text{BE}(G_i, \Theta_i)$. However, notice that if we introduce assumptions $\Theta_i$ that restrict the other agents to $\llbracket \Theta_i \rrbracket$, then reasoning in terms of best effort under $\Theta_i$ is unnecessary, because the same restriction can be used directly to seek $\sigma_i \in \text{ENF}(G_i,\Theta_i)$.
% \end{definition}

Note that if a strategy $\sigma_i$ is best-effort for $G_i$ against all strategies $\sigma_{-i} \in \Sigma_{-i}$, $\sigma_{e} \in \llbracket \Gamma \rrbracket$, then it is a winning strategy.
Therefore, we can observe that all winning strategies are best-effort, but not vice versa.

Given a set of strategies (or strategy profiles), we are interested in the traces that are \emph{consistent with} that set, i.e., traces that can be generated by some strategy (profile) in the set. Notice that our notion of consistency should not be confused with the typical notion of consistency of a process model, interpreted as the possibility of producing at least one accepting execution.
% Strategy sets induce sets of possible executions. \lnote{Consistency w.r.t. a given strategy or strategy profile (not consistency in the usual sense)}.

\begin{definition}[Consistent Trace]
For a set of strategy profiles $\Sigma$, %$\upsilon = \sigma_1, \ldots, \sigma_n$, 
we denote the set of traces $t \in (2^{\mathfrak{A}})^\omega$ that are consistent with $\Sigma$ by 
$\text{CNS}(\Sigma)
\;=\;
\{\, \textit{play}(\upsilon,\sigma_e)  \ |\ 
\exists \upsilon \in \Sigma,\
\exists \sigma_e \in \llbracket \Gamma \rrbracket
\}$.
Similarly, for one agent $\mathcal{A}_i$, we denote the set of traces $t \in (2^{\mathfrak{A}})^\omega$ that are consistent with its set of strategies $\Sigma_i$  by
$\text{CNS}_i(\Sigma_i)
\;=\;
\{\,
\textit{play}(\sigma_i,\sigma_{-i},\sigma_e)
\ |\ 
\exists \sigma_i \in \Sigma_i,\
\exists \sigma_{-i},\
\sigma_e \in \llbracket \Gamma \rrbracket
\,\}$.

\end{definition}

Intuitively, $\text{CNS}$ turns a set of strategies into the set of all infinite traces that could happen when one of those strategies is used~\cite{KR2025-69}.

\section{Possible Settings of Agentic BPM}
\label{sec:settings}
In this section, we formally introduce the three different settings of agentic BPM (cf. Figure~\ref{fig:tree}) and how sets of agent strategies can be obtained, respectively, using the concepts defined before. 
We describe each setting in terms of its input and output and refer to our motivating example process to provide some intuitions.
%

% \begin{table}[t]
% \centering
% \setlength\tabcolsep{6pt}
% \caption{Comparison of agentic BPM control settings. Notice that the three settings are uniquely characterized by decision-making (centralized vs. decentralized) and the type of agent assumptions.}
% \begin{tabular}{l|rrr}
% \toprule
%  & \textbf{Decision-making} & \textbf{Assumption} & \textbf{References} \\
% \midrule
% Setting 1 & centralized & none & \cite{DBLP:conf/aips/AminofGMR19,DBLP:conf/kr/AminofGLMR21} \\
% % Setting 2 & centralized & n/a & \cite{DBLP:conf/kr/AminofGLMR21} \\
% % \hdashline
% % Setting 3 & decentralized  & explicit & \cite{KR2025-69}\\
% Setting 2 & decentralized  & explicit & \cite{KR2025-69,DBLP:conf/ijcai/AminofGR21} \\
% % \hdashline
% % Setting 4o & decentralized & BE & no & \\
% Setting 3 & decentralized & implicit & \cite{DBLP:conf/stacs/Berwanger07} \\
% \bottomrule
% \end{tabular}
% \label{tab:settings}
% \end{table}

\subsection{Centrally Controlled Agentic BPM}
In the centrally controlled setting, a single controller determines the
strategies of all agents, i.e., the controller prescribes the strategy profile $\upsilon = \sigma_1,\ldots,\sigma_n$.
% The objective of the controller is to specify a strategy profile that guarantees satisfaction of the joint goal under all admissible environment behaviors.

\begin{description}[style=unboxed,leftmargin=0cm]
    \item[Input.] 
    Let $(\Gamma,\mathcal{A},\mathcal{G})$ be an agentic process with agents $\mathcal{A}_1,\ldots,\mathcal{A}_n$ and individual goals  $G_1,\ldots,G_n$, and let $G = G_1 \land \cdots \land G_n$ denote the conjunction of individual goals.

    \item[Output.] 
    We are interested in the set of winning strategy profiles $\Sigma = \text{ENF}(G)$. If the global goal is not enforceable, i.e., $\text{ENF}(G)=\emptyset$, we instead consider the set of best-effort strategy profiles $\Sigma = \text{BE}(G)$, which maximize goal satisfaction relative to all other strategy profiles under $\Gamma$.
\end{description}

% In the centrally controlled setting, a single controller determines the
% behavior of all agents by selecting a strategy profile
% $\upsilon = \sigma_1,\ldots,\sigma_n$.
% The environment remains external and chooses valuations for the fluent
% variables according to a strategy $\sigma_e \in \llbracket \Gamma \rrbracket$.

% The objective of the controller is to specify a strategy profile that
% guarantees satisfaction of the joint goal under all admissible
% environment behaviors.
% Hence, we are interested in the set of winning strategy profiles
% $\Sigma = \text{ENF}(G)$.
% If the global goal is not enforceable, i.e.,
% $\text{ENF}(G)=\emptyset$, we instead consider the set of
% best-effort strategy profiles $\Sigma = \text{BE}(G)$.
% These profiles maximize goal satisfaction relative to all other
% strategy profiles under $\Gamma$.

\mypar{Running example.} 
Computing $\text{ENF}(G)$ for the travel expense process yields all strategy profiles that guarantee the three agent goals, regardless of whether the environment sets the documentation quality to good or bad. 
In one such winning profile, the controller prescribes that the traveler submits the request through the customer-facing route whenever documentation is good, and through the internal route otherwise. The controller further prescribes that the expense handler then approves well-documented customer-facing requests and requests resubmission otherwise, while the manager approves all internal requests.
% In one such winning profile, the controller may specify that the traveler submits the request through the customer-facing route whenever the documentation is good, and switches to the internal route otherwise. The expense handler approves well-documented customer-facing requests and asks for resubmission otherwise, while the manager approves all internal requests. 
% Hence, the controller coordinates the three agents so that, for every possible choice of documentation quality by the environment, all goals are satisfied.

% \lnote{IN WORDS}
% \[
% \begin{aligned}
% \upsilon_1:\quad
%     &\sigma_1 = (DT \land DG \rightarrow AD)\ \land\ (DT \land \lnot DG \rightarrow RE),\\
%     &\sigma_2 = (AD \land DG \rightarrow AR)\ \land\ (AD \land \lnot DG \rightarrow SE),\\
%     &\sigma_3 = RE \rightarrow AR\\
% \end{aligned}
% \]

% Intuitively, $\upsilon_1$ prescribes that the traveler chooses the customer-facing option when documentation is good and the internal option otherwise; consequently, customer-facing trips are always approved by the expense handler and internal trips by the manager.

\subsection{Decentralized Agentic BPM With Explicit Assumptions}
In contrast to the centralized setting, there is no controller in decentralized agentic BPM.
Instead, each agent $\mathcal{A}_i$ autonomously selects its own strategy
$\sigma_i$.

\begin{description}[style=unboxed,leftmargin=0cm]
    \item[Input.] 
    Let $(\Gamma,\mathcal{A},\mathcal{G})$ be an agentic process with agents $\mathcal{A}_1,\ldots,\mathcal{A}_n$ and individual goals $G_1,\ldots,G_n$. Further, let $\Sigma_{-i}$ be the set of strategy profiles of all agents except $\mathcal{A}_i$.
    In this particular setting, we let $\Sigma_{-i}$ be the assumptions that agent $\mathcal{A}_i$ has about its peers.

    \begin{definition}[Agent Assumptions]
    Let $(\Gamma,\mathcal{A},\mathcal{G})$ be an agentic process and $i \in \{1,\dots,n\}$.  
    An \emph{assumption} for agent $\mathcal{A}_i$ is 
    $\Theta_i = \Sigma_{-i} = \Sigma_1 \times \dots \times\Sigma_{i-1} \times \Sigma_{i+1} \times \ldots \times \Sigma_n,$
    where each $\Sigma_j$ is a set of strategies for agent $\mathcal{A}_j$.
    Intuitively, the set of strategies $\Theta_i$ represents the agent's view of the behaviors it might face from the other agents.
    In other words, $\Theta_i$ is the set of strategies of the other agents:
    $\Theta_i
    \;=\;
    \{
    \sigma_{-i} \,|\,
    \forall j \neq i:\ \sigma_j \in \Sigma_j 
    \}$.
    
    % $\llbracket\Theta_i\rrbracket$ the set of strategies that are compliant with $\Theta_i$, i.e.,  
    % \[
    % \llbracket \Theta_i \rrbracket
    % \;=\;
    % \Big\{
    % \sigma_{-i} \,\Big|\,
    % \forall j \neq i:\ \sigma_j \in \Sigma_j 
    % \Big\}.
    % \]
    \end{definition}

    \item[Output.] 
    For each agent $\mathcal{A}_i$, we are interested in the set of winning (or, alternatively, best-effort) strategies that satisfy its individual goal $G_i$ against all behaviors of other agents consistent with $ \Theta_i$.
    \[
    \begin{aligned}
    \Sigma_i &= \text{ENF}_i(G_i,\Sigma_{-i}') 
        && \text{where } \Sigma_{-i}' = \Theta_i \\
    (\Sigma_i &= \text{BE}_i(G_i,\Sigma_{-i}') 
        && \text{where } \Sigma_{-i}' =  \Theta_i )
    \end{aligned}
    \]
    
\end{description}

\mypar{Running example.} 
For the travel expense process, suppose that the traveler assumes that the expense handler behaves as in the centralized setting, namely by approving well-documented customer-facing requests and asking for resubmission when the documentation is poor, while the manager is assumed to deny every request that reaches the internal branch. 
That is, $\Theta_1 = (\Sigma_2, \Sigma_3)$, where  $\Sigma_2= \text{ENF}_2((AD \land DG \rightarrow AR)\ \land\ (AD \land \lnot DG \rightarrow SE))$ and $\Sigma_3 = \text{ENF}_3(RE \rightarrow DR)$.
Under these assumptions, the traveler can guarantee its own goal only by always choosing the customer-facing route and never the internal one. Intuitively, once the internal branch is assumed to be blocked by the manager, it is no longer a viable option for a winning strategy.

\subsection{Decentralized Agentic BPM With Implicit Assumptions}
An interesting variant of the above setting is obtained by not fixing the assumptions $\Theta_1,\ldots,\Theta_n$.

\begin{description}[style=unboxed,leftmargin=0cm]
    \item[Input.] 
    Let $(\Gamma,\mathcal{A},\mathcal{G})$ be an agentic process with agents $\mathcal{A}_1,\ldots,\mathcal{A}_n$ and individual goals $G_1,\ldots,G_n$. Further, let $\Sigma_{-i}$ be the set of strategy profiles of all agents except $\mathcal{A}_i$.

    \item[Output.]
    For each agent $\mathcal{A}_i$, we are interested in computing the set of winning strategies and simultaneously the assumption for them to win
    \[
    \Sigma_i = \text{ENF}_i(G_i,\Sigma_{-i}) \qquad i \in \{1,\dots,n\}.
    \]
    Formally, what we get is a set of equations whose variables are $\Sigma_i$. Such equations can have an arbitrary number of solutions (possibly zero), and we need some Pareto-optimal criteria to choose among them. We are interested in computing the maximal set of strategies for each $\Sigma_i$ to give maximal freedom to each agent $\mathcal{A}_i$. However, as we increase the freedom of an agent $\mathcal{A}_i$, we might actually decrease the freedom of another agent $\mathcal{A}_j$.
    % In other words, one can enlarge one of the sets (give more freedom to one agent) by restricting the other (restricting the freedom of the other agents). In other words, we want to maximize the number of strategies in each of the sets.
    % Still, e.g., maximizing $\Sigma_1$ may, e.g., require shrinking $\Sigma_2$ and $\Sigma_3$.
    So, how to choose a suitable Pareto-optimal criterion remains in general unclear, requiring further work.
    
    Interestingly, if we focus on best-effort, we can provide a single maximal solution to these equations.
    That is, the set of equations 
    \[
    \Sigma_i = \text{BE}_i(G_i,\Sigma_{-i}) \qquad i \in \{1,\dots,n\}
    \]
    admits a single maximal solution, where each agent has the maximal freedom in best-effort satisfying its goal, assuming that each agent does the same.

\end{description}

\mypar{Running example.}
Returning to the travel expense process, now suppose that all three agents reason in a best-effort manner, each choosing an undominated strategy under the assumption that the others do the same. In the resulting maximal solution, the traveler distinguishes between the two documentation cases: well-documented trips are submitted as customer-facing, whereas poorly documented ones are redirected to the internal route. The intuition is that a best-effort expense handler approves strong customer-facing requests, but has no incentive to keep weak ones alive if denying them already secures its own goal better than asking for resubmission. Anticipating this, the traveler avoids sending poorly documented requests through a branch that would likely be closed off, and instead chooses the internal route, where approval may still remain possible. In this sense, the traveler acts best-effort by preserving every remaining opportunity to achieve its goal under the others' best-effort behavior.

\
%%%%%%%%%%%%%%%%%%%%%%%

\section{Foundational Problems of Agentic Processes}
\label{sec:services}

The strategy-set semantics introduced in Section~\ref{sec:formal} provides a uniform representation of agentic processes across the different settings presented in Section~\ref{sec:settings}.
Beyond serving as a descriptive semantics, the representation of strategy sets enables a range of fundamental analysis tasks concerning the feasibility, construction, correctness, and monitoring of agentic processes.
In this section, we characterize four such problems that arise naturally in agentic BPM. Throughout all problems, we let $(\Gamma,\mathcal{A},\mathcal{G})$ be an agentic process, and let $\llbracket \Gamma \rrbracket$ be the set of admissible environment strategies. \\

%\subsection{Realizability}
%\label{sec:realizability}
\noindent \textbf{Realizability} determines whether there exists a strategy (or strategy profile) that enforces the intended goal(s) under the process constraints.
We next instantiate the check for each of the settings from Section~\ref{sec:settings}. 

Notice, however, that we only focus on enforcing strategy sets because the corresponding best-effort strategy sets are guaranteed to be non-empty (cf. \cite{DBLP:conf/ijcai/AminofGR21}).

    \noindent
    \textbf{Setting 1 (centralized).} The process is \emph{realizable} iff a strategy can enforce the joint goal:
        $\Sigma = \text{ENF}(G_1 \land \cdots \land G_n)\ \neq\ \emptyset$.
    
    \noindent
    \textbf{Setting 2 (decentralized, explicit assumptions).}
     We say that the goal $G_i$ of agent $\mathcal{A}_i$ is realizable under $\Theta_i$ iff
      $\Sigma_i = \text{ENF}_i(G_i,\Sigma'_{-i})\ \neq\ \emptyset$, where $\Sigma'_{-i} = \Theta_i$. \\

%\subsection{Synthesis}
%\label{sec:synthesis}
%While realizability is the decision problem of whether the relevant strategy sets are non-empty, 
\noindent \textbf{Synthesis}
is the constructive counterpart to realizability, as we compute an \emph{explicit} strategy (or strategy profile) witnessing realizability.

\noindent\textbf{Setting 1 (centralized).}
The synthesis task is to return a finite-state strategy profile
$\upsilon \in \text{ENF}(G_1\land\cdots\land G_n)$ if it exists; otherwise,
return a finite-state profile $\upsilon \in \text{BE}(G_1\land\cdots\land G_n)$.

\noindent\textbf{Setting 2 (decentralized, explicit assumptions).}
For each agent $\mathcal{A}_i$, synthesize a finite-state strategy
$\sigma_i \in \text{ENF}_i(G_i,\Theta_i)$ if it exists; otherwise, return a best-effort strategy for each agent $\sigma_i \in \text{BE}_i(G_i,\Theta_i)$.

\noindent\textbf{Setting 3 (decentralized, implicit assumptions).}
For each agent $\mathcal{A}_i$, synthesize a finite-state strategy
$\sigma_i \in \text{BE}_i(G_i, \Sigma_{-i})$. \\

%\subsection{Model-Checking}
%\label{sec:modelchecking}
\noindent \textbf{Model-checking} verifies that, \emph{regardless of how agents choose within their allowed strategy spaces}, every resulting trace satisfies an additional temporal requirement, such as eventually reaching request approval as in our example process.
Let $\varphi$ be an LTL formula over $\mathfrak{A}$, $(\Gamma,\mathcal{A},\mathcal{G})$ be an agentic process and write $\llbracket \varphi \rrbracket \subseteq (2^{\mathfrak{A}})^\omega$ for the set of traces satisfying $\varphi$.

\noindent\textbf{Setting 1 (centralized).}
Let $\Sigma = \text{ENF}(G)$ be a set of strategy profiles, decide whether
$\text{CNS}(\Sigma) \subseteq \llbracket \varphi \rrbracket$.

\noindent\textbf{Settings 2--3 (decentralized).}
Let $\Sigma_i = \text{ENF}_i(G_i,\Theta_i)$ (resp. $\Sigma_i = \text{BE}_i(G_i, \Sigma_{-i})$ for Setting 3).
Decide whether
$\bigcap_{i=1}^n \text{CNS}_i(\Sigma_i)\ \subseteq\ \llbracket \varphi \rrbracket$. \\

%\subsection{Guardrailing}
%\label{sec:guardrail}
\noindent \textbf{Guardrailing} detects when the observed execution can no longer be explained by \emph{any} allowed strategy of an agent (or strategy profile), intuitively %, and can therefore be attributed as a deviation from the prescribed strategy space. 
%One may consider this as conformance checking between agent executions and strategies (which essentially serve as guardrails),
lifting \emph{frames} in the sense of declarative process specifications and their corresponding conformance checks from the trace to the strategy level.
In other words, if we have a history that violates $\text{CNS}_i(\Sigma_i)$, we know that agent $\mathcal{A}_i$ is not adopting a strategy in $\Sigma_i$ and so $\mathcal{A}_i$ is responsible for the breach of $\Sigma_i$.
In these cases, guardrailing allows us to raise a red flag and indicate that $\mathcal{A}_i$ is not acting correctly.
Formally, let a \emph{history} be a finite prefix of a trace.
For a set of traces $T \subseteq (2^{\mathfrak{A}})^\omega$, define its prefix-closure
$\mathrm{Pref}(T)
=
\{
h \in (2^{\mathfrak{A}})^* \,|\,
\exists t \in T : h \text{ is a prefix of } t
\}$.
Intuitively, $\mathrm{Pref}(T)$ contains exactly the histories that can be extended to a full trace in $T$.

\noindent
\noindent\textbf{Setting 1 (centralized).}
Let $\Sigma$ be the controller's allowed strategy-profile space (e.g., $\Sigma=\text{ENF}(G)$ if non-empty, otherwise $\Sigma=\text{BE}(G)$ as in Section~\ref{sec:settings}).
A history $h$ is \emph{compliant} iff
$h \in \mathrm{Pref}\big(\text{CNS}(\Sigma)\big)$.
Otherwise, $h$ is a \emph{guardrail violation}, and the controller is responsible for the breach (since no allowed profile in $\Sigma$ can explain the observed behavior under any admissible environment strategy).

Moreover, if $h$ is \emph{minimal} w.r.t.\ violation (all strict prefixes of $h$ are compliant), then the violation is pinpointed to the last joint action prescribed by the controller (i.e., the last valuation of $X_1 \cup \cdots \cup X_n$ along $h$), because that is the step at which the set of $\Sigma$-compatible profiles becomes empty.

\noindent\textbf{Settings 2--3: decentralized guardrailing.}
In our decentralized settings, guardrailing is performed per agent.
Fix an agent $\mathcal{A}_i$ and its allowed strategy space $\Sigma_i$ (Setting~2: $\Sigma_i=\text{ENF}_i(G_i,\Theta_i)$ or $\text{BE}_i(G_i,\Theta_i)$; Setting~3: $\Sigma_i=\text{BE}_i(G_i,\Sigma_{-i})$).
A history $h$ is \emph{$i$-compliant} iff
$h \in \mathrm{Pref}\big(\text{CNS}_i(\Sigma_i)\big)$.

If $h \notin \mathrm{Pref}(\text{CNS}_i(\Sigma_i))$, then there exists \emph{no} strategy $\sigma_i \in \Sigma_i$ that remains compatible with the observed execution prefix, and thus $\mathcal{A}_i$ is responsible for violating its prescribed strategy space.
As in the centralized case, if $h$ is a minimal violating history for $\mathcal{A}_i$, then the last move of $\mathcal{A}_i$ (i.e., the last valuation of $X_i$ in $h$) is the cause of the violation, because immediately before that move there still existed a non-empty subset of strategies in $\Sigma_i$ consistent with the history, but afterwards none remain \cite{KR2025-69}.

\section{Instantiation in LTL}
\label{sec:ltl_instantiation}

We now instantiate our set-based framework of the previous chapters with LTL specifications. 
Concretely, we take the operational process specification $\Gamma$ and all agent goals $G_1, \ldots, G_n$ to be given as LTL formulas $\varphi$ over the global propositional alphabet $Z \cup X_1 \cup \ldots X_n$, where $X_i$ denotes the action variables controlled by agent $\mathcal{A}_i$, and $Z$ denotes fluents controlled by the environment.

Consistently with our earlier notation $\llbracket \Gamma \rrbracket$ (as a set of strategies; see Definition~\ref{def:gamma}), in the LTL instantiation we can express with 
$\llbracket \Gamma \rrbracket = \text{ENF}_e(\Gamma)$
the set of environment strategies that enforce the LTL process specification $\Gamma$. 
If we consider LTL goals $\varphi$ and assumptions $\Sigma_{-i}$ in order to restrict the strategies of the other agents, we obtain:
$\text{ENF}_i(\varphi, \Sigma_{-i}) = \{\sigma_i \; | \; \forall \sigma_e \in \llbracket \Gamma \rrbracket.\ \forall \sigma_{-i} \in \Sigma_{-i}.\ \textit{play}(\sigma_i, \sigma_{-i}, \sigma_e)\models \varphi \}$.
We abbreviate the case where $\Sigma_{-i}$ is unspecified as
$\text{ENF}_i(\varphi) = \{\sigma_i \; | \; \forall \sigma_e \in \llbracket \Gamma \rrbracket.\ \forall \sigma_{-i} .\ \textit{play}(\sigma_i, \sigma_{-i}, \sigma_e)\models \varphi \}$. 

Notice that we can let assumptions themselves be specified by LTL expressions that denote strategy sets. Concretely, for each agent $\mathcal{A}_i$, we denote the set of strategies $\Sigma_i$ through strategy expressions $\alpha_i$ with the following syntax~\cite{KR2025-69}:
\[
\alpha_i = \text{ENF}_i(\varphi) \;\Big| \; \text{ENF}_i(\varphi, \Theta_i), \qquad \Theta_i = (\alpha_1, \ldots, \alpha_{i-1}, \alpha_{i+1}, \ldots, \alpha_n), 
\]

\noindent so that $\Theta_i(j) = \alpha_j$ describes which strategy set agent $j \neq i$ is assumed to choose from. Furthermore, notice that our assumption expression $\Theta_i$ can be nested (and even mutually referential) through these expressions. With a little abuse of notation, we are going to use $\Theta_i$ and $\alpha_i$ instead of the set of strategies that they denote $\Sigma_{-i}$ and $\Sigma_i$.

$\text{CNS}_{-i}(\Theta_i)$ denotes the set of traces consistent with at least one strategy profile allowed by the assumption $\Theta_i$. Crucially, following \cite{KR2025-69}, we can compile assumptions away by an implication over traces:

\[
\text{ENF}_i(\varphi, \Theta_i) = \text{ENF}_i(\text{CNS}_{-i}(\Theta_i) \Rightarrow \varphi).
\]
Most importantly, $\text{CNS}_{-i}(\Theta_i)$ is itself an $\omega$-regular property whose automaton is obtained from that of $\Theta_i$ (recursively) via a linear operation.
We can reduce synthesis under assumptions to standard synthesis for an LTL-definable (or automata-definable) trace property.

% All foundational problems, introduced in Section~\ref{sec:services} are \textsc{2EXPTIME} in the worst-case.
% The reason is that, in every problem, the complexity is dominated by computational the construction of the automaton that captures the \emph{set of strategies}, and the induced \emph{game arena} on top of it. 

\begin{theorem}[Setting 1]
Let $\Gamma$ be the operational process specification and $G_1,\ldots,G_n$ be the goals of agents $\mathcal{A}_1,\ldots,\mathcal{A}_n$ in setting 1, all expressed in LTL.
\begin{compactitem}[$\bullet$]
    \item Checking realizability and synthesizing a strategy profile is 2EXPTIME-complete in $\Gamma$ and $G_1,\ldots,G_n$ both for winning and best-effort.

    \item Let $\varphi$ be a global LTL property of the process. Then, model-checking for $\varphi$ is 2EXPTIME-complete in $\Gamma$ and $G_1,\ldots,G_n$, and PSPACE-complete in $\varphi$.

    \item Let $\varphi$ be an LTL property. Then, guardrailing the profile for $\varphi$ is 2EXPTIME-complete in $\Gamma$ and $G_1,\ldots,G_n$, and PSPACE-complete in $\varphi$. Moreover, once constructed, guardrailing is \emph{linear} in the length of the observed history.
    
\end{compactitem}
\end{theorem}

% \lnote{Add proof sketch (ICAPS and IJCAI21)}
% \cite{DBLP:conf/aips/AminofGMR19} ICAPS for synthesis and \cite{DBLP:conf/ijcai/AminofGR21} for best-effort.

\begin{proof}[Proof sketch]
% In Setting~1, the whole strategy profile
% $\upsilon=(\sigma_1,\ldots,\sigma_n)$ is controlled centrally, while the
% environment ranges over $\llbracket\Gamma\rrbracket$.
% Hence 
To solve all these problems, we need to build the game-arena needed for synthesis for $\Gamma \Rightarrow G$ where $G = G_1 \land \cdots \land G_n$, which is 2EXPTIME in the worst-case. On the other hand, hardness comes from LTL synthesis.
The result for realizability and synthesis is a direct consequence of this oberservation (cf.~\cite{DBLP:conf/aips/AminofGMR19} for winning and ~\cite{DBLP:conf/ijcai/AminofGR21} for best-effort solutions).

For model checking, let $\Sigma$ be either $\mathrm{ENF}(G)$ or $\mathrm{BE}(G)$. We obtain: 
$\mathrm{CNS}(\Sigma)\subseteq \llbracket\varphi\rrbracket
\text{ iff }
\mathrm{CNS}(\Sigma)\cap \llbracket\neg\varphi\rrbracket=\emptyset$.
Constructing the automaton for $\mathrm{CNS}(\Sigma)$ where $\Sigma= \text{ENF}(G) \text{ or BE}(G)$ is linear in the size of the game-arena for $\Sigma$. Considering that building the nondetermnistic B\"uchi automaton for $\neg\varphi$ can be done on-the-fly while checking for emptiness~\cite{BaierK08} we get the results above. 

For guardrailing we say a history $h$ is compliant iff
$h\in \mathrm{Pref}(\mathrm{CNS}(\Sigma))$. Therefore, guardrail violation reduces
to bad-prefix detection for $\mathrm{CNS}(\Sigma)$, yielding the same bounds as for model checking. Once the monitor automaton for bad prefixes is built, it processes the observed history in a single pass, hence guardrailing is linear in the history length.
\end{proof}

\begin{theorem}[Setting 2]
Let $\Gamma$ be the operational process specification and $G_1,\ldots,G_n$ be the goals, and $\Theta_1,\ldots,\Theta_n$ be the assumptions according to the syntax above for agents $\mathcal{A}_1,\ldots,\mathcal{A}_n$ in setting 2, all expressed in LTL.
\begin{compactitem}[$\bullet$]
    \item Checking realizability and synthesis for an agent $\mathcal{A}_i$ strategy $\sigma_i$ is 2EXPTIME-complete in $\Gamma$, $G_i$, and $\Theta_i$ both for winning and best-effort.

    \item Let $\varphi$ be a global LTL property of the process. Then, model-checking for $\varphi$ is 2EXPTIME-complete in $\Gamma$, $G_1,\ldots,G_n$, and $\Theta_1,\ldots,\Theta_n$, and PSPACE-complete in $\varphi$.

    \item Let $\varphi_i$ be an LTL property. Then, guardrailing  $\mathcal{A}_i$ for $\varphi_i$ is 2EXPTIME-complete in $\Gamma$, $G_i$, and $\Theta_i$ and PSPACE-complete in $\varphi_i$. Moreover, once constructed, guardrailing is linear in the length of the observed history.
    
\end{compactitem}
\end{theorem}
% \lnote{Add proof sketch (KR25)}
% \cite{KR2025-69}
\begin{proof}[Proof sketch]
The arguments are similar to Setting~1, except that we now focus on single agents $\mathcal{A}_i$ and we need to handle assumption $\Theta_i$.
A key result in~\cite{KR2025-69} shows that 
$\mathrm{ENF}_i(G_i,\Theta_i)
=
\mathrm{ENF}_i(\mathrm{CNS}_{-i}(\Theta_i)\Rightarrow G_i)$. This actually applies through the recursive structure of the assumption expression $\Theta_i$, each subexpression is compiled bottom-up into a deterministic
$\omega$-automaton for its $\mathrm{CNS}$ language.
So we can rely on standard techniques for synthesis under assumptions, i.e., ~\cite{DBLP:conf/aips/AminofGMR19} for winning and ~\cite{DBLP:conf/ijcai/AminofGR21} for best-effort solutions.

For model checking we need to further observe that given $\Sigma_i$ the allowed strategy set of each
agent (winning or best-effort under $\Theta_i$), we build automata for each $\mathrm{CNS}_i(\Sigma_i)$ and intersect them, 
thus getting
$\bigcap_{i=1}^n \mathrm{CNS}_i(\Sigma_i)\subseteq \llbracket\varphi\rrbracket$.
Given these observations we get the results above. 
\end{proof}

\begin{theorem}[Setting 3]
Let $\Gamma$ be the operational process specification and $G_1,\ldots,G_n$ be the goals of agents $\mathcal{A}_1,\ldots,\mathcal{A}_n$ in setting 3, all expressed in LTL.
\begin{compactitem}[$\bullet$]
    \item Synthesizing a best-effort strategy $\sigma_i$ for agent $\mathcal{A}_i$ is 2EXPTIME-complete in $\Gamma$ and $G_1,\ldots,G_n$.

    \item Let $\varphi$ be a global LTL property of the process. Then, model-checking for $\varphi$ is 2EXPTIME-complete in $\Gamma$ and $G_1,\ldots,G_n$, and PSPACE-complete in $\varphi$.

    \item Let $\varphi_i$ be an LTL property. Then, guardrailing  $\mathcal{A}_i$ for $\varphi_i$ is 2EXPTIME-complete in $\Gamma$ and $G_1,\ldots,G_n$, and PSPACE-complete in $\varphi_i$. Moreover, once constructed, guardrailing is linear in the length of the history.
    
\end{compactitem}
\end{theorem}

% \lnote{Add proof sketch (Dietmar)}
\begin{proof}[Proof sketch]
Unlike Setting~2, the assumptions are not given explicitly, but are defined as the greatest solutions of the equations $\Sigma_i=\mathrm{BE}_i(G_i,\Sigma_{-i})$. Hence Setting~3 amounts to computing the maximal fixed point of mutual
best-effort strategies~\cite{DBLP:conf/stacs/Berwanger07}.
As in the previous settings, we translate $\Gamma$ and the goals
$G_1,\ldots,G_n$ into the corresponding game-arena of doubly
exponential size. Over this arena, the sets $\Sigma_i$ are obtained by
iterated elimination of dominated strategies until a fixed point is reached, in polynomial steps in the number of agents and size of the game-arena. 
Once the fixed point is computed, we can apply the same constructions of Setting~2, thus getting the results above. 
\end{proof}

%%%%%% New version from Giuseppe:
\mypar{Discussion on implementability.}
Complexity bounds refer to worst-case instances of the problem, often not representative of average instances, nor of instances found in practice. This has a twofold implication. On the one hand, even problems falling within intractable complexity classes such as PSPACE and 2EXPTIME (matching different problems arising in this paper) often come with effective algorithms providing strong performance on average instances. On the other hand, distinct problems falling within the same complexity class may lead to more or less effective implementations.
A prime example substantiating these observations is the LTL$_{f}$ synthesis problem~\cite{DBLP:conf/ijcai/GiacomoV15}: it has the same complexity (2EXPTIME-complete) as LTL synthesis over infinite traces, but it is widely regarded as success story in AI due to the strong practical scalability of its reference implementations~\cite{DBLP:conf/wia/DuretLutzZPGV25}.
While a detailed investigation of implementation feasibility for the
problems considered here is beyond our scope, we note that effective
techniques can be developed by exploiting results on efficient synthesis.

\section{Conclusion}
\label{sec:discussion}
%%%%%%%%%%%%%%%%%%%%%%%
We have provided formal foundations for agentic BPM systems where multiple agents collectively execute a process, while aiming at achieving their own goal. 
The notions introduced here can be instrumental to increase safety and trustworthiness of agentic BPM systems employing generative AI-based agents~\cite{10.1007/978-3-032-02936-2_3}. Notably, our work is not constrained to any specific agent technology; it may also be used to represent and reason about \emph{human} agents in business processes. 

Our formal foundations of agentic BPM can be utilized by future work across all three dimensions of the BPM community.
In the \emph{formal} dimension, a notable open question is the definition of a Pareto-optimality notion that fairly balances freedom of action between agents, as discussed in Section~\ref{sec:settings}. Also of interest is extending our approach to include quantitative aspects (like costs and incentives, as studied in~\cite{DBLP:conf/bpm/HeindelW20}), and to extend the \emph{agent system mining} framework for discovering agent specifications from event data~\cite{DBLP:conf/bpm/TourPKS23} with the possibility of discovering strategies.
In the \emph{engineering} dimension, applied work may integrate our symbolic agentic BPM foundations with sub-symbolic (learning) agents and multi-agent systems. 
Finally, in the \emph{management} dimension, researchers may develop new frameworks that advance agentic BPM in practice.

\section*{Acknowledgments}
This work was supported in part by the UKRI Erlangen AI Hub on Mathematical and Computational Foundations of AI (No. EP/Y028872/1). 
TK was partially supported by the Wallenberg AI, Autonomous Systems and Software Program (WASP) funded by the Knut and Alice Wallenberg Foundation.

\bibliographystyle{splncs04}
\bibliography{references}
\end{document}